\documentclass[letterpaper, 10 pt, conference]{ieeeconf}  

\IEEEoverridecommandlockouts                              

\usepackage{epsfig} 
\usepackage{amsmath} 
\usepackage{amssymb}  
\usepackage{threeparttable}
\usepackage{xcolor}
\usepackage{fancyhdr}
\usepackage{float}
\usepackage{colortbl}
\usepackage{dcolumn}
\usepackage{siunitx}
\usepackage{diagbox}
\usepackage[colorlinks=true,linkcolor=blue,citecolor=blue,urlcolor=blue]{hyperref}

\title{\LARGE \bf
Lyapunov Function-guided Reinforcement Learning for Flight Control}

\author{Yifei Li$^{1}$ and Erik-Jan van Kampen$^{2}$
\thanks{$^{1}$Yifei Li is with the Faculty of Aerospace Engineering, Delft University of Technology, 2629HS, Delft, The Netherlands
        {\tt\small Y.Li-34@tudelft.nl}}%
\thanks{$^{2}$Erik-Jan van Kampen is with the Faculty of Aerospace Engineering, Delft University of Technology,
        Delft, 2629HS, The Netherlands
        {\tt\small E.vanKampen@tudelft.nl }}%
}

\begin{document}

\maketitle

\begin{abstract}
A cascaded online learning flight control system has been developed and enhanced with respect to action smoothness. In this paper, we investigate the convergence performance of the control system, characterized by the increment of a Lyapunov function candidate. The derivation of this metric accounts for discretization errors and state prediction errors introduced by the incremental model. Comparative results are presented through flight control simulations.

\end{abstract}
\emph{\textbf{Index Terms}}-\textbf{reinforcement learning, action smoothness, flight control, filter.}

\section{INTRODUCTION}

The performance of flight control systems is evaluated with multiple criteria. One important criterion is the ability to converge to the equilibrium points. This convergence has been analyzed and configured for Linear Time-Invariant (LTI) systems by the pole placement approach, which aims to arrange the locations of poles to desirable places in the \textit{s}-plane\cite{bib1,bib2,bib3,bib4,bib5,bib6,bib7}. The convergence speed generally increases when the poles are located further to the left in the \textit{s}-plane, i.e., when they have more negative real parts. Examples of this approach are algebraic pole placement (APP) \cite{bib3}, continuous pole placement (CPP) \cite{bib4,bib7} and partial pole placement (PPP) \cite{bib6}. However, this approach is inapplicable to nonlinear or time-varying systems, as computing accurate pole locations requires solving nonlinear equations, which is often impractical. 

Since 1892, the Lyapunov's methods have been proposed to analyze stability of nonlinear and time-varying systems \cite{bib8}. This original research proposed two analysis methods: Lyapunov's first and second methods. In the framework of Lyapunov's second method, the first derivative of a Lyapunov function $V(x)$ is utilized to quantify convergence speed of states. This also benefits the process of control designs that achieve state regulation or state tracking. As such, the principle of control design lies in achieving the continuing decrease of the Lyapunov function, i.e. $\dot{V}(x)\leq 0$. Consequently, system states are ensured to asymptotically converge to equilibrium points. In previous literature, this idea has been leveraged in model-based control designs to appropriately assign the structure and parameters of control laws, such as Nonlinear Dynamic Inversion (NDI), Backstepping (BS) and Sliding Mode Control (SMC). On the other hand, the extend of the decrease of $V(x)$ also implies the state convergence speed. As such, the convergence speed can be controlled by tuning the values of $\dot{V}(x)$. 

The improvement on convergence performance can also be considered in data-driven control methods. For example, the data-driven parameter adaptation laws are designed to guarantee decrease of a Lyapunov function \cite{bib9,bib10,bib11}. This design further leads to an asymptotic convergence of system states. In infinite-horizon optimal control problem, the optimal control law that minimizes a value function has been proved to provide state-convergent performance \cite{bib12,bib13,bib14}. The convergence speed can be adjusted by the weights between the quadratic states and quadratic actions in an one-step cost \cite{bib13}. Moreover, the process of policy learning by RL methods implicitly improves the convergence ability granted by an initial policy. This is a result of minimizing a cost function associated with states and state-feedback control, which relates the minimization of the cost function and convergence speed of states. However, this cost function does not explicitly measure the convergence speed.

The purpose of this chapter is to design a measure that \textit{explicitly} represents the convergence performance of system states. This measure will be further used as a loss item to guide policy training. This learning method will be verified in a cascaded online learning flight control system. The remainder of this chapter is organized as follows. Section \ref{section_model} formulates the tracking error dynamics for angle of attack and pitch rate. Section \ref{section_Equilibrium_Point_Analysis} analyzes the equilibrium point of tracking error dynamics. Section \ref{section_Lyapunov Function-guided IHDP} introduces the Lyapunov function-guided IHDP method, which employs a convergence measure. Section \ref{section_simulation} provides simulation results on the cascaded online learning flight control. Section \ref{section_conclusion} concludes this chapter.

\section{Model}\label{section_model}

This section introduces the longitudinal dynamical model of aerial vehicles. The discrete-time model is then derived based on the Euler method for convenient flight control design.

\subsection{Dynamics}
The nonlinear dynamical equations of aerial vehicles are given as

\begin{equation} \label{aerialvehiclemodel}
\begin{aligned}
    \dot{\alpha} &= (\frac{fgQS}{WV})\cos(\alpha)[\phi_{z}(\alpha)+b_{z}\delta] + q\\
    \dot{q} &= (\frac{fQSd}{I_{yy}})[\phi_{m}(\alpha)+b_{m}\delta]
\end{aligned}
\end{equation}\\
where $\alpha, q$ are angle-of-attack, pitch rate, $\delta$ is the control surface deflection. The aerodynamic coefficients are approximately computed by $b_{z}=-0.034, b_{m} = -0.206$, and 

\begin{equation}
\begin{aligned}
    &\phi_{z}(\alpha) = 0.000103\alpha^{3}-0.00945\alpha|\alpha|-0.170\alpha\\
    &\phi_{m}(\alpha) = 0.000215\alpha^{3} - 0.0195\alpha|\alpha| - 0.051\alpha
\end{aligned}
\end{equation}

These approximations of $b_{z},b_{m},\phi_{z}(\alpha),\phi_{m}(\alpha)$ hold for $\alpha$ in the range of $\pm 20$ degrees. The physical coefficients are provided in Table \ref{physcial_coefficient_aerial_vehicle}. In addition, the actuator dynamics are considered as a first order model with the time constant $0.005s$. The rate limit is 600 deg/s, and a control surface deflection limit is $\pm 20$ degrees.

\begin{table}[htbp]
  \centering
  \caption{Physical parameters (adapted from \cite{bib20})}
  \begin{threeparttable}
    \begin{tabular}{lll}
    \hline
    Notations & Definition & Value\\
    \hline
    $g$ & acceleration of gravity & 9.815 m/s$^{2}$\\
    $W$ & weight & 204.3 kg\\
    $V$ & speed & 947.715 m/s\\
    $I_{yy}$ & pitch moment of inertia & 247.438 kg$\cdot$ m$^{2}$\\
    $f$ & radians to degrees & 180/$\pi$\\
    $Q$ & dynamic pressure & 29969.861 kg/m$^{2}$\\
    $S$ & reference area & 0.041 m$^{2}$\\
    $d$ & reference diameter & 0.229 m\\
    \hline
    \end{tabular}
 \end{threeparttable}
 \label{physcial_coefficient_aerial_vehicle}
\end{table}

\subsection{Tracking dynamics}
The aerial vehicle dynamics \ref{aerialvehiclemodel} associated to angle-of-attack and pitch rate are provided, the equations of tracking errors can be formulated as



\begin{equation} \label{alpha_error_dynamics}
\begin{aligned}
    \dot{e}_{1} &= (\frac{fgQS}{WV})\cos(\alpha)[\phi_{z}(\alpha)+b_{z}\delta] + q - \dot{\alpha}_{\text{ref}}\\
    \dot{e}_{2} &= (\frac{fQSd}{I_{yy}})[\phi_{m}(\alpha)+b_{m}\delta]-\dot{q}_{\text{ref}}
\end{aligned}
\end{equation}\\
where $e_{1}=\alpha-\alpha_{\text{ref}},e_{2}=q-q_{\text{ref}}$ are defined as tracking errors for angle-of-attack and pitch rate references denoted as $\alpha_{\text{ref}}$, $q_{\text{ref}}$.

\section{Equilibrium Point Analysis}\label{section_Equilibrium_Point_Analysis}
The definition of the equilibrium point for a continuous-time system is given as follows.

\textbf{Definition 1.} \label{definition_equilibrium_point}
    \cite{bib15}
    The point $x_{e}\in\mathbb{R}^{m}$ is an equilibrium point for the differential equation $\dot{x}=f(t,x)$, if $f(t,x_{e})=0$ for all $t$. 

Denote $\alpha_{o}$ as the equilibrium point of angle-of-attack, which is regarded as an intermediate variable to compute equilibrium point $e_{1o}$ according to $e_{1o}=\alpha_{o}-\alpha_{\text{ref}}$. According to Definition \ref{definition_equilibrium_point}, the set of all equilibrium points of angle-of-attack is given by


\begin{equation} \label{equilibrium_point}
\begin{aligned}
    \mathcal{D} = \Big\{\alpha_{o}\Big|\big(\frac{fgQS}{WV}\big)\cos(\alpha_{o})[\phi_{z}(\alpha_{o})+b_{z}\delta]\\
    + W^{\vartheta_{1}}(\alpha_{o}-\alpha_{\text{ref}},\alpha_{o}) - \dot{\alpha}_{\text{ref}}=0\Big\}
\end{aligned}
\end{equation}


According to \ref{equilibrium_point}, the equilibrium point is moving over time due to the time-varying $\alpha_{\text{ref}}, \dot{\alpha}_{\text{ref}}$ and time-varying actor parameter set $\vartheta_{1}$ in the process of policy learning. The design of actor input as $e_{1},\alpha$ enables learning a control law that cancels internal dynamics and provides proportional control simultaneously. As the internal dynamics is canceled, the equilibrium point $e_{1o}$ gets close to $e_{1o}=0$.


The state $e_{1}$ will converges to the equilibrium point $e_{1o}$ if the closed-loop system is stable. However, the equilibrium point $e_{1o}$ does not lie in the $e_{1o}=0$ due to the error of canceling internal dynamics. This is verified as follows.\\
\textbf{Verification of the claim $e_{1}=0$ is not an equilibrium point}

Rewrite Equation \ref{alpha_error_dynamics} as

\begin{equation} \label{alpha_error_dynamics_detailed}
\begin{aligned}
    \dot{e}_{1} =& (\frac{fgQS}{WV})\cos(\alpha)[\phi_{z}(\alpha)+b_{z}\delta] + q - \dot{x}_{\text{ref}}\\
    =& (\frac{fgQS}{WV})\cos(e_{1}+\alpha_{\text{ref}})[\phi_{z}(e_{1}+\alpha_{\text{ref}})+b_{z}\delta] + q - \dot{\alpha}_{\text{ref}}\\
    =& (\frac{fgQS}{WV})\cos(e_{1}+\alpha_{\text{ref}})[0.000103(e_{1}+\alpha_{\text{ref}})^{3}\\    &-0.00945(e_{1}+\alpha_{\text{ref}})|e_{1}+\alpha_{\text{ref}}|-0.170(e_{1}+\alpha_{\text{ref}})\\
    &+b_{z}\delta] + q - \dot{\alpha}_{\text{ref}}
\end{aligned}
\end{equation}

By property of cosine operator and cube operator, one has
\begin{equation} \label{cosine_cube}
\begin{aligned}
    \cos(\alpha+\beta)=\cos\alpha\cos\beta-\sin\alpha\sin\beta\\
    (a+b)^{3} = a^{3}+b^{3}+3a^{2}b+3ab^{2}    
\end{aligned}
\end{equation}

Substitute Equations \ref{cosine_cube} into Equation \ref{alpha_error_dynamics_detailed}:

\begin{equation} \label{alpha_error_dynamics_detailed}
\begin{aligned}
    \dot{e}_{1} =& (\frac{fgQS}{WV})\cos(e_{1}+\alpha_{\text{ref}})[0.000103(e_{1}+\alpha_{\text{ref}})^{3}\\    &-0.00945(e_{1}+\alpha_{\text{ref}})|e_{1}+\alpha_{\text{ref}}|-0.170(e_{1}+\alpha_{\text{ref}})\\
    &+b_{z}\delta] + q - \dot{\alpha}_{\text{ref}}\\
    =& (\frac{fgQS}{WV})[\cos(e_{1})\cos(\alpha_{\text{ref}})-\sin(e_{1})\sin(\alpha_{\text{ref}})]\\&[0.000103(e^{3}_{1}+\alpha^{3}_{\text{ref}}+3e^{2}_{1}\alpha_{\text{ref}}+3e_{1}\alpha^{2}_{\text{ref}})\\    
    &-0.00945\text{sign}(e_{1}+\alpha)(e^{2}_{1}+\alpha^{2}_{\text{ref}}+2e_{1}\alpha_{\text{ref}})\\  &-0.170(e_{1}+\alpha_{\text{ref}})+b_{z}\delta] + q - \dot{\alpha}_{\text{ref}}
\end{aligned}
\end{equation}

By making $e_{1}=0$, one has
\begin{equation} 
\begin{aligned}
    \dot{e}_{1}|_{e_{1}=0} 
    =& (\frac{fgQS}{WV})\cos(\alpha_{\text{ref}})(0.000103\alpha^{3}_{\text{ref}}\\
    &-0.00945\alpha_{\text{ref}}|\alpha_{\text{ref}}|    -0.170\alpha_{\text{ref}}+b_{z}\delta)\\
    &+ q - \dot{\alpha}_{\text{ref}}\neq 0
\end{aligned}
\end{equation}

The fact $\dot{e}_{1}|_{e_{1}}\neq0$ indicates $e_{1}=0$ is not an equilibrium point. 

\subsubsection{Explaination of $e_{1}$ divergence }
The previous analysis shows that the tracking error $e_{1}$ may converge to an equilibrium point $e_{1}\neq 0$. This results into the phenomenon that $e_{1}$ moves away from $e_{1}=0$. By RL, the process of minimizing value function $\hat{V}$ is achieved by increasing proportional gain that reduces effects of internal dynamics. This process implicitly leads to a result of $\hat{V}$ decreases, even this property is not considered in reward function. However, the decrease of $\hat{V}$ is an expected performance of the closed-loop system, that enables $e_{1}\rightarrow 0$, which should be emphasized in the process of policy training.   
  
\section{Lyapunov Function-guided IHDP}\label{section_Lyapunov Function-guided IHDP}
\subsection{Discrete-time Lyapunov function increment}
The discrete-time Lyapunov function increment $\hat{V}(x_{t+1})-\hat{V}(x_{t})$ is a measure to optimize the closed-loop system performance. This is inspired from the asymptotic stability condition. This condition is based on a continuous state-action space so that it can not be directly used in the discrete state-action space. This motivates the derivation of a discrete-time measure.

A deterministic discrete-time nonlinear system is given as

\begin{equation}\label{nonlinearsystem}	x_{t+1}=f(x_{t},u_{t}), t\in \mathbb{N}
\end{equation}\\
where $f: \mathbb{R}^{n}\times \mathbb{R}^{m} \rightarrow \mathbb{R}^{n}$ is a smooth nonlinear function associated with state vector $x_{t}$ and input vector $u_{t}$. $n,m$ are positive integers denoting the dimensions of the state and control spaces. $t$ represents the discrete-time index. $\mathbb{N}$ represents the set of non-negative integers.

\textbf{Assumption 1.}\label{Lipschitz_continuity_assumption_f}
    The dynamics $f(\cdot)$ in \ref{nonlinearsystem} is Lipschitz continuous with respect to the 1-norm.

\textbf{Definition 2.}
    The point $x_{e}\in\mathbb{R}^{m}$ is an equilibrium point for the difference equation \ref{nonlinearsystem} if $f(x_{e},u_{t})=x_{e}$ for $t=0,1,2,\cdots$. 

\textbf{Lemma 1.} \cite{bib16} Using Assumption \ref{Lipschitz_continuity_assumption_f}, let $[x]_{\tau}$ be discretized state of $x_{\tau}$, $\mathcal{X}_{\tau}$ be a discretization of state space $\mathcal{X}$ such that $\Vert x-[x]_{\tau} \Vert_{1}\leq \tau$ for all $\textbf{x}\in\mathcal{X}$. Then, for all $x\in\mathcal{X}$, we have
\begin{equation}
\begin{aligned}
    &|v(\mu_{n-1}([z]_{\tau}))-v([x]_{\tau})-(v(f(z))-v(z))|\\
    \leq& L_{v}\beta_{n}\sigma_{n-1}([z]_{\tau})+L_{\Delta
     v}\tau
\end{aligned}
\end{equation}\\
where $z=(x,\pi(x)),[z]_{\tau}=([x]_{\tau},\pi([x]_{\tau}))$, $L_{\Delta v}=(L_{v}L_{f}(L_{\pi}+1)+L_{v})$.

\textbf{Proof.}    
Let $z=(x,\pi(x)), [z]_{\tau} = ([x]_{\tau},\pi([x]_{\tau}))$, and $f=f_{n-1},\sigma = \sigma_{n-1}$.

\begin{equation}\label{LyapunovStability}
\begin{aligned}
        &|v(\mu_{n-1}([z]_{\tau}))-v([x]_{\tau}) - (v(f(z))-v(x))|\\
       =&|v(\mu_{n-1}([z_{\tau}))-v([x]_{\tau}) - v(f(z) + v(x))|\\
       =&|v(\mu_{n-1}([z]_{\tau})) - v(f([z]_{\tau})) +  v(f([z]_{\tau}))\\
       &- v(f(z) + v(x))-v([x]_{\tau}))|\\ 
       =&|v(\mu_{n-1}([z]_{\tau})) - v(f([z]_{\tau}))| +|  v(f([z]_{\tau})) - v(f(z)|\\
       &+ |v(x))-v([x]_{\tau}))|\\ 
       \leq & L_{v}\Vert \mu_{n-1}([z]_{\tau}) - f([z]_{\tau})\Vert_{1} + L_{v}\Vert f([z]_{\tau}) - f(z)\Vert_{1}\\
       &+ L_{v}\Vert x - [x]_{\tau}\Vert_{1}\\
       \leq & L_{v}(\delta+\epsilon)([z]_{\tau})+ L_{v}L_{f}\Vert [z]_{\tau} - z\Vert_{1} + L_{v}\Vert x - [x]_{\tau}\Vert_{1}
\end{aligned}
\end{equation}

According to definition of discretization, one has
\begin{equation} \label{discretizationmethod}
\begin{aligned}
    \Vert z - [z]_{\tau} \Vert_{1} &= \Vert x - [x]_{\tau} \Vert + \Vert \pi(x) -\pi([x]_{\tau}) \Vert_{1}\\
    &\leq \tau + L_{\pi}\Vert x - [x]_{\tau}\Vert_{1}\\
    &\leq (L_{\pi}+ 1)\tau
\end{aligned}
\end{equation}

Substitute \ref{discretizationmethod} into \ref{LyapunovStability}:
\begin{equation}\label{lema3}
\begin{aligned}
      &|v(\mu([z]_{\tau})) - v([x]_{\tau}) - (v(f(z))-v(x))| \\
      \leq& L_{v}(\delta+\epsilon) ([z]_{\tau})
      + [L_{v}L_{f}(1+ L_{\pi})+L_{v}]\tau
\end{aligned}    
\end{equation}

By absolute value inequality, Inequality \ref{lema3} is rewritten as

\begin{equation} \label{noabsolutevalue}
\begin{aligned}
      &-L_{v}(\delta + \epsilon) ([z]_{\tau})
      - [L_{v}L_{f}(1+ L_{\pi})+L_{v}]\tau \\
      \leq&   v(\mu([z]_{\tau})) - v([x]_{\tau}) - (v(f(z))-v(x)) \\
      \leq& L_{v}(\delta + \epsilon) ([z]_{\tau})
      + [L_{v}L_{f}(1+ L_{\pi})+L_{v}]\tau
\end{aligned}    
\end{equation}

Recall the left side of \ref{noabsolutevalue}:
\begin{equation} 
\begin{aligned}
      & -L_{v}(\delta + \epsilon) ([z]_{\tau})
      - [L_{v}L_{f}(1+ L_{\pi})+L_{v}]\tau\\
      \leq & v(\mu([z]_{\tau})) - v([x]_{\tau}) - (v(f(z))-v(x))
\end{aligned}    
\end{equation}
which is rewritten as
\begin{equation}
\begin{aligned}
      &v(f(z))-v(x) \leq v(\mu([z]_{\tau})) - v([x]_{\tau})+L_{v}(\delta + \epsilon) ([z]_{\tau})\\
      +& [L_{v}L_{f}(1+ L_{\pi})+L_{v}]\tau
\end{aligned}    
\end{equation}

Recall the decrease condition of $v(\cdot)$:
\begin{equation} \label{decreasecondition}
\begin{aligned}
    v(f(z))-v(x) \leq 0
\end{aligned}    
\end{equation}

To make Inequality \ref{decreasecondition} hold, a sufficient but unnecessary condition is 
\begin{equation}\label{mucondition}
\begin{aligned}
      &v(\mu([z]_{\tau})) - v([x]_{\tau})+L_{v}(\delta + \epsilon) ([z]_{\tau})\\
      &+ [L_{v}L_{f}(1+ L_{\pi})+L_{v}]\tau \leq 0 
\end{aligned}    
\end{equation}

Substitute $u([z]_{\tau}) = \mu([z]_{\tau}) + L_{v}(\delta + \epsilon) ([z]_{\tau})$ into \ref{mucondition}:
\begin{equation}
\begin{aligned}
      v(u([z]_{\tau})) - v([x]_{\tau})
      + [L_{v}L_{f}(1+ L_{\pi})+L_{v}]\tau \leq 0 
\end{aligned}    
\end{equation}

Because
\begin{equation}\label{absineq}
\begin{aligned}
      &|v(\mu([z]_{\tau})) - v([x]_{\tau})| -| (v(f(z))-v(x)) |  \\
      \leq&  |v(\mu([z]_{\tau})) - v([x]_{\tau}) - (v(f(z))-v(x)) |
\end{aligned}    
\end{equation}

Substitute \ref{lema3} into \ref{absineq}:
\begin{equation}\label{absineq}
\begin{aligned}
      &|v(\mu([z]_{\tau})) - v([x]_{\tau})| -| (v(f(z))-v(x)) | \\
      \leq & L_{v}\beta_{n} \sigma ([z]_{\tau})
      + [L_{v}L_{f}(1+ L_{\pi})+L_{v}]\tau
\end{aligned}    
\end{equation}

Then
\begin{equation}\label{absineq}
\begin{aligned}
      &| (v(f(z))-v(x)) | 
      \geq  |v(\mu([z]_{\tau})) - v([x]_{\tau})| \\
      &-L_{v}\beta_{n} \sigma ([z]_{\tau}) - [L_{v}L_{f}(1+ L_{\pi})+L_{v}]\tau
\end{aligned}    
\end{equation}

By decrease condition, one has
\begin{equation}\label{decrease_condition_conclusion}
\begin{aligned}
      &|v(\mu([z]_{\tau})) - v([x]_{\tau})|-L_{v}\beta_{n} \sigma ([z]_{\tau})\\
      -& [L_{v}L_{f}(1+ L_{\pi})+L_{v}]\tau \geq 0\\
      &|v(\mu([z]_{\tau})) - v([x]_{\tau})|-L_{v}\beta_{n} \sigma ([z]_{\tau})\\
      \geq& [L_{v}L_{f}(1+ L_{\pi})+L_{v}]\tau \\
\end{aligned}    
\end{equation}

Inequality \ref{decrease_condition_conclusion} provides a practical condition for decrease condition $v(x_{t+1})-v(x_{t})\leq 0$, i.e.
\begin{equation}\label{practicaldecrase}
    \mu_{n}(x,u)<v(x)-L_{\Delta_{v}}\tau
\end{equation}

\subsection{Modification on IHDP}
In 2016, Incremental-model-based Heuristic Dynamic Programming (IHDP) has been developed by Zhou \cite{bib17}. In this framework, an incremental model is adopted to approximate the nonlinear system, which provides reduced computation compared to a model network utilized in HDP \cite{bib18}. To apply IHDP for flight control design, the one-step cost function is commonly designed as a quadratic function of tracking errors and actions to achieve the performance balance between control precision and control effort. On the other hand, the minimization of a quadratic function also improves convergence of tracking error. An disadvantage of this approach is that stability is degraded by various approximation errors. To optimize the convergence performance of the closed-loop system in an explicitly pattern, the convergence metric can be used to guide the actor training:



By considering the stability measure, we modify the policy optimization:


\begin{equation} \label{safe_policy_opt}
\begin{aligned}
        \pi_{n} =& \underset{\pi_{\theta}\in\prod_{L}}{\arg\min}\sum_{x\in\mathcal{X}_{\tau}} r(x,\pi_{\theta}(x))+\gamma J_{\pi_{\theta}}(f(x,\pi_{\theta}(x))\\
        &+\lambda(u_{n}(x,\pi_{\theta}(x))-v(x)+L_{\Delta_{v}}\tau)
\end{aligned}
\end{equation}\\
where $\lambda$ is a Lagrangian multiplier. The prior model $\mu_{n-1}(x,\pi_{\vartheta}(x))$ and its Lyapunov function upper bound $u_{n}(x,\pi_{\vartheta}(x))$ is used.

To further simplify the optimization objective, we set $\lambda$ as a manually specified coefficient, and since $v(x)$ does not propagate gradient to $\vartheta$, and $L_{\Delta_{v}}\tau$ is a constant, they can be ignored. Equation \ref{safe_policy_opt} is then simplified as
\begin{equation}\label{practical_policy_optimization}
\begin{aligned}
        \pi_{n} =& \underset{\pi_{\theta}\in\prod_{L}}{\arg\min}\sum_{x\in\mathcal{X}_{\tau}}r(x,\pi_{\vartheta}(x))\\
        &+\gamma J_{\pi_{\vartheta}}(f(x,\pi_{\vartheta}(x))
    +\lambda(v(x,\pi_{\vartheta}(x))
\end{aligned}
\end{equation}



\section{Simulation}\label{section_simulation}

\subsection{Reward shaping}

The one-step cost function for tracking control tasks is usually designed as a quadratic function associated with tracking error and action \cite{bib19}. This design enables a performance trade-off between tracking error and control effort. In a cascaded online learning flight control system, the one-step cost functions are separately designed for each subsystem.

The one-step cost for the higher-level agent is

\begin{equation}
\begin{aligned}  \label{agent_1_r1} 
    c_{1(t)} = \hat{e}_{\alpha(t+1)}^{2} + aq_{\text{ref}(t)}^{2}
\end{aligned}
\end{equation}\\
where $a>0$ is the weight of quadratic pitch rate reference. 

The one-step cost for the lower-level agent is
\begin{equation}
\begin{aligned}  \label{agent_2_r2} 
    c_{2(t)} = \hat{e}_{q(t+1)}^{2} + b\delta_{t}^{2}\end{aligned}
\end{equation}\\
where $b>0$ is the weight of quadratic control surface deflection.

\subsection{Results and discussion}

This subsection provides extended simulations, aimed at investigating the convergence of the control system. The basic settings are consistent. The aerial vehicle dynamics are seen in Equation \ref{aerialvehiclemodel}. The online temporal smoothness and a low-pass filter are used to improve action smoothness. Additionally, the convergence metrics are employed in policy optimization of both higher-level and lower-level agents. The reference signal is defined as $\alpha_{\text{ref}} = 10^{\circ}\sin(\frac{2\pi}{T}t), T=10s$. The sampling time and control period are both set to 0.001s. The remaining parameters are seen in Table \ref{Hyperparameters_of_RL_agents}.

\begin{table}[htbp]
  \centering
  \caption{Hyperparameters of RL agents}
  \begin{threeparttable}
    \begin{tabular}{l|l|l}
    \hline
    Parameter & Higher-level agent & Lower-level agent \\
    \hline
    critic learning rate $\eta_{C_{1}},\eta_{C_{2}}$& 0.1 & 0.1\\
    actor learning rate $\eta_{A_{1}},\eta_{A_{2}}$ & $5\times 10^{-7}$ & $10^{-7}$\\
    discount factor $\gamma$ & 0.6 & 0.6\\
    delay factor $\tau$ & 1 & 1\\
    forgetting factor $\alpha$ & 0.99 & 0.99\\
    policy iteration number at $t$ & 3 & 3\\
    hidden layer size & 7 & 7 \\
    critic hidden layer activation function & tanh & tanh \\
    critic output layer activation function & abs &  abs \\
    actor activation function & tanh & tanh \\
    optimizer &Adam &Adam\\
    weight on the quadratic error & 1  & 1 \\ 
    weight on the quadratic action $a,b$& $5\times 10^{-6}$  & $10^{-5}$\\ 
    weight on the smoothness loss $\rho$ & $9.3\times 10^{-3}$ & $10^{-5}$\\ 
    threshold for critic loss & $5\times 10^{-5}$ & $10^{-4}$\\ 
    Maximum update steps for critic and actor & 50 & 50\\ 
    \hline 
    \end{tabular}
 \end{threeparttable}
 \label{Hyperparameters_of_RL_agents}
\end{table}

\subsubsection{Higher-level agent}
\begin{figure}[htbp]
    \centering
    \includegraphics[width=1.0\linewidth]{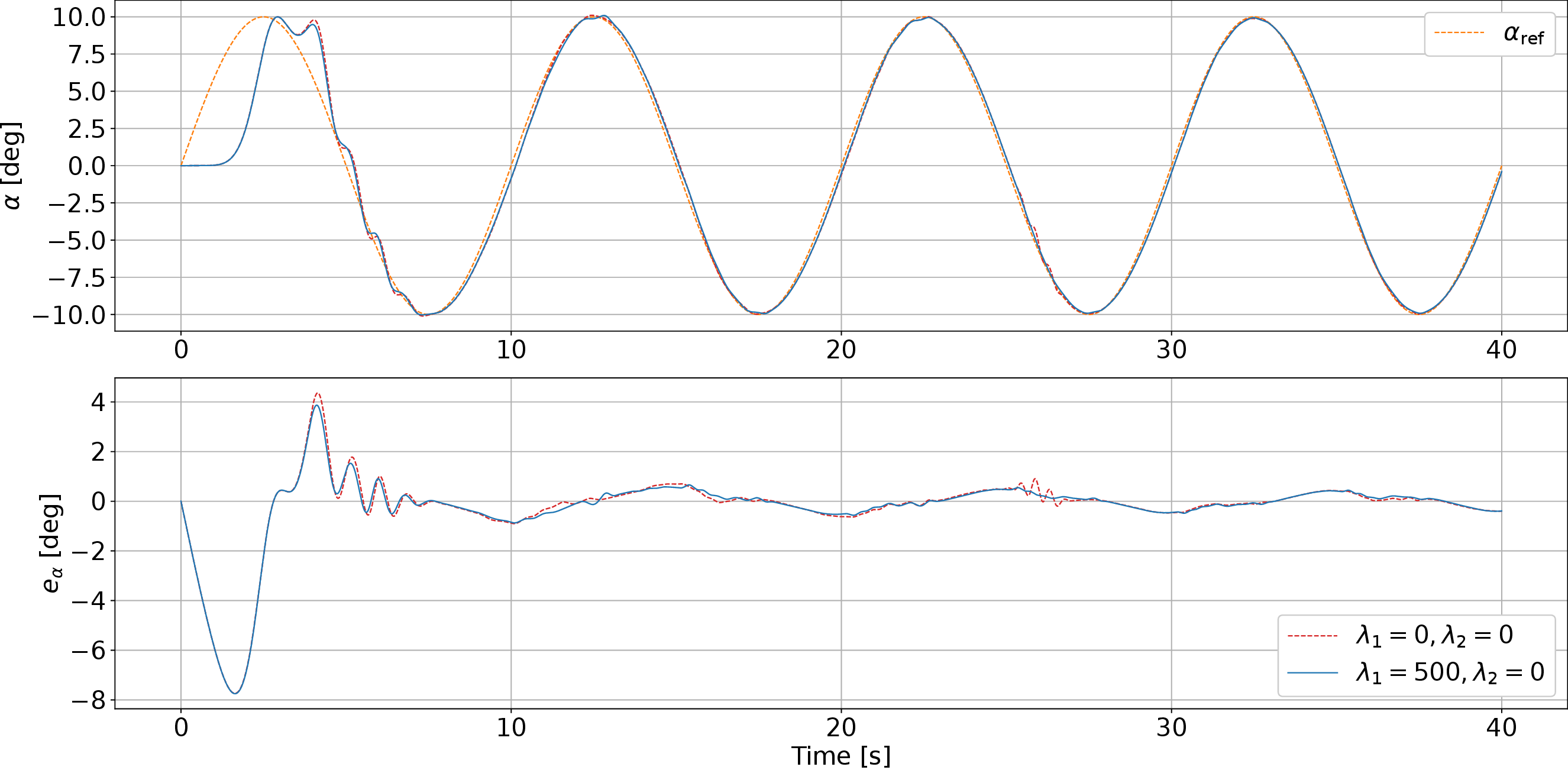}
    \caption{Comparison of angle-of-attack tracking between IHDP and Lyapunov function-guided IHDP used by the higher-level agent.} \label{alpha_tracking_Cha4}
\end{figure}

Figure \ref{alpha_tracking_Cha4} compares angle-of-attack tracking of two flight control systems. The first control system uses IHDP in the higher-level agent, while the second control system uses IHDP with a convergence metric. Successful angle-of-attack trackings are observed for both two control systems, while using a convergence metric ($\lambda_{1}=500$) slightly improves the convergence of tracking error $e_{\alpha}$ in the time period 5-10s. $\lambda_{1}$ is the weight of the convergence metric for the higher-level agent's actor. 

\begin{figure}[htbp]
    \centering 
    \includegraphics[width=1.0\linewidth]{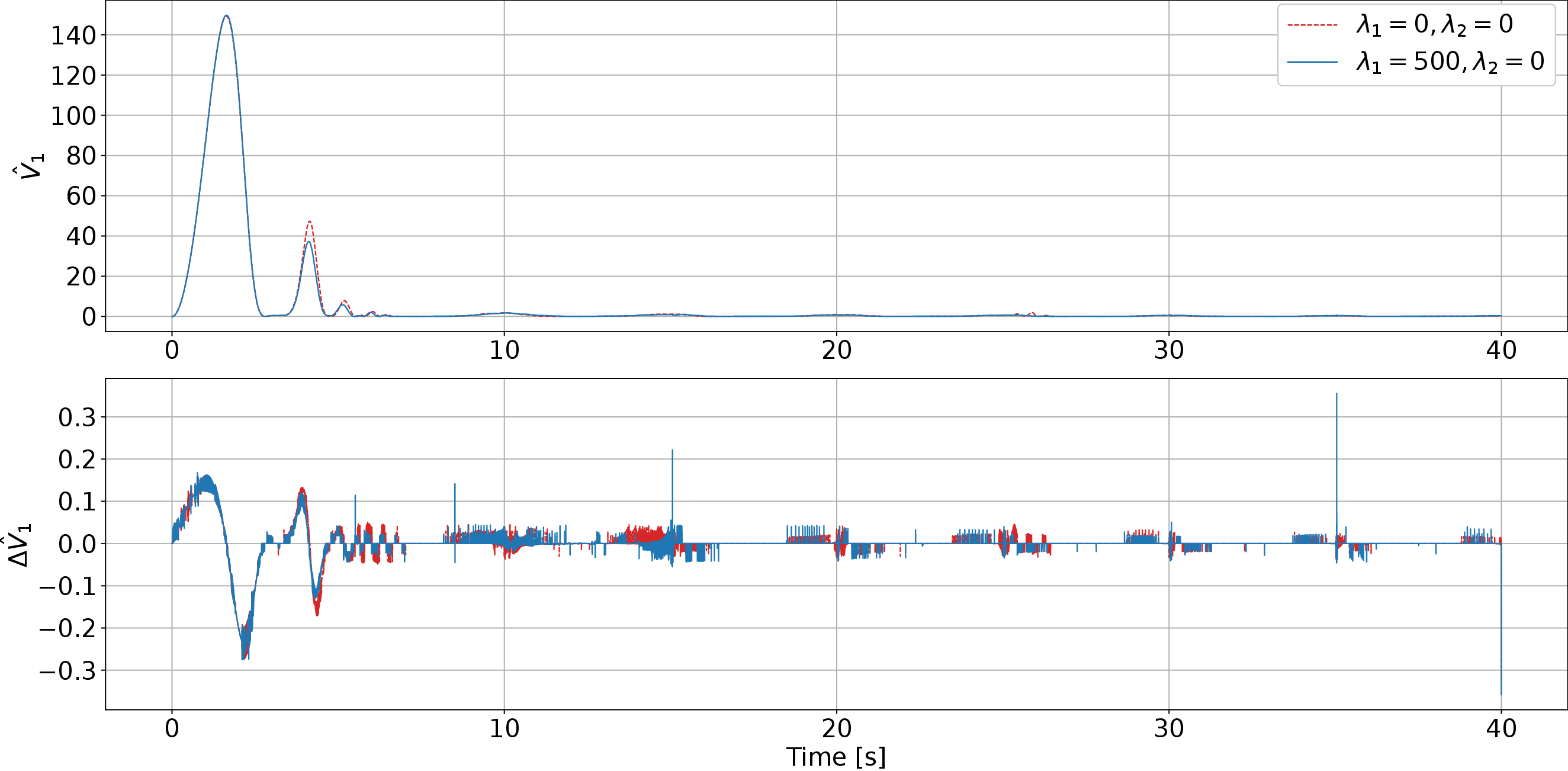}
    \caption{Comparison of critic output and its increment for the higher-level agent.} \label{alpha_Lyapunov_Cha4}
\end{figure}

Figure \ref{alpha_Lyapunov_Cha4} compares the Lyapunov function candidate $\hat{V}_{1}$. The increase of $\hat{V}_{1}$ during the initial phase of policy learning indicates that the state $e_{\alpha}$ are leaving the expected equilibrium point $e_{\alpha}=0$. According to Equation \ref{alpha_error_dynamics_detailed}, the term $\dot{\alpha}_{\text{ref}}$ affects the rate of error $\dot{e}_{1}$. In the policy learning phase 0-10s, the control gains do not provide sufficient control effects to offset $\dot{\alpha}_{\text{ref}}$. Define the increment of Lyapunov function as $\Delta \hat{V}_{1(t)} = \hat{V}_{1(t+1)}-\hat{V}_{1(t)}$. The comparison of this measure in the second subplot of Figure \ref{alpha_Lyapunov_Cha4} shows that using a convergence metric slightly strengthens the decrease of $\Delta \hat{V}_{1(t)}$.

\subsubsection{Lower-level agent}

\begin{figure}[htbp]
    \centering
    \includegraphics[width=1.0\linewidth]{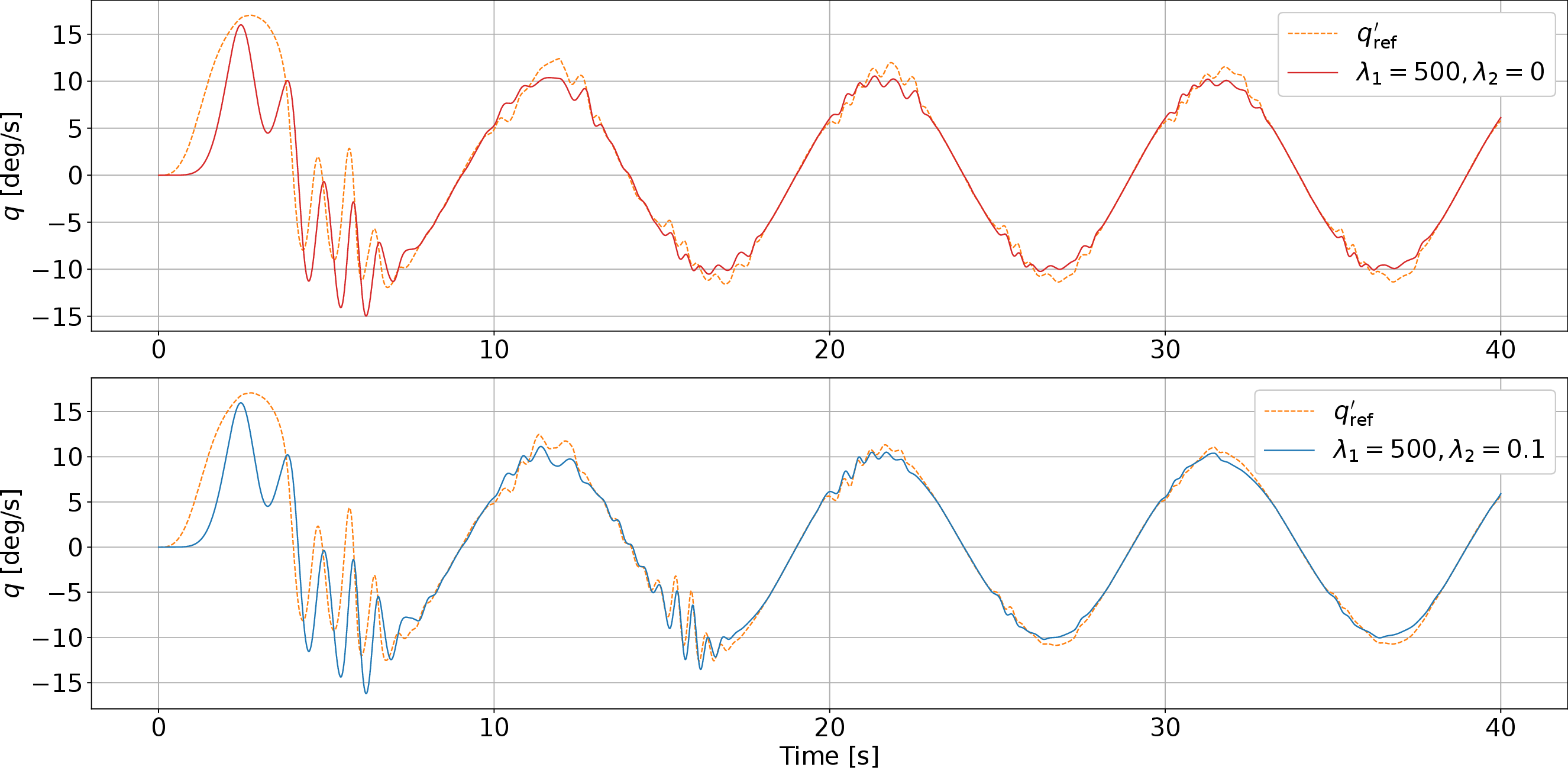}
    \caption{Comparison of pitch rate tracking between IHDP and Lyapunov function-guided IHDP used by the lower-level agent.} \label{q_tracking_Cha4}
\end{figure}

\begin{figure}[htbp]
    \centering
    \includegraphics[width=1.0\linewidth]{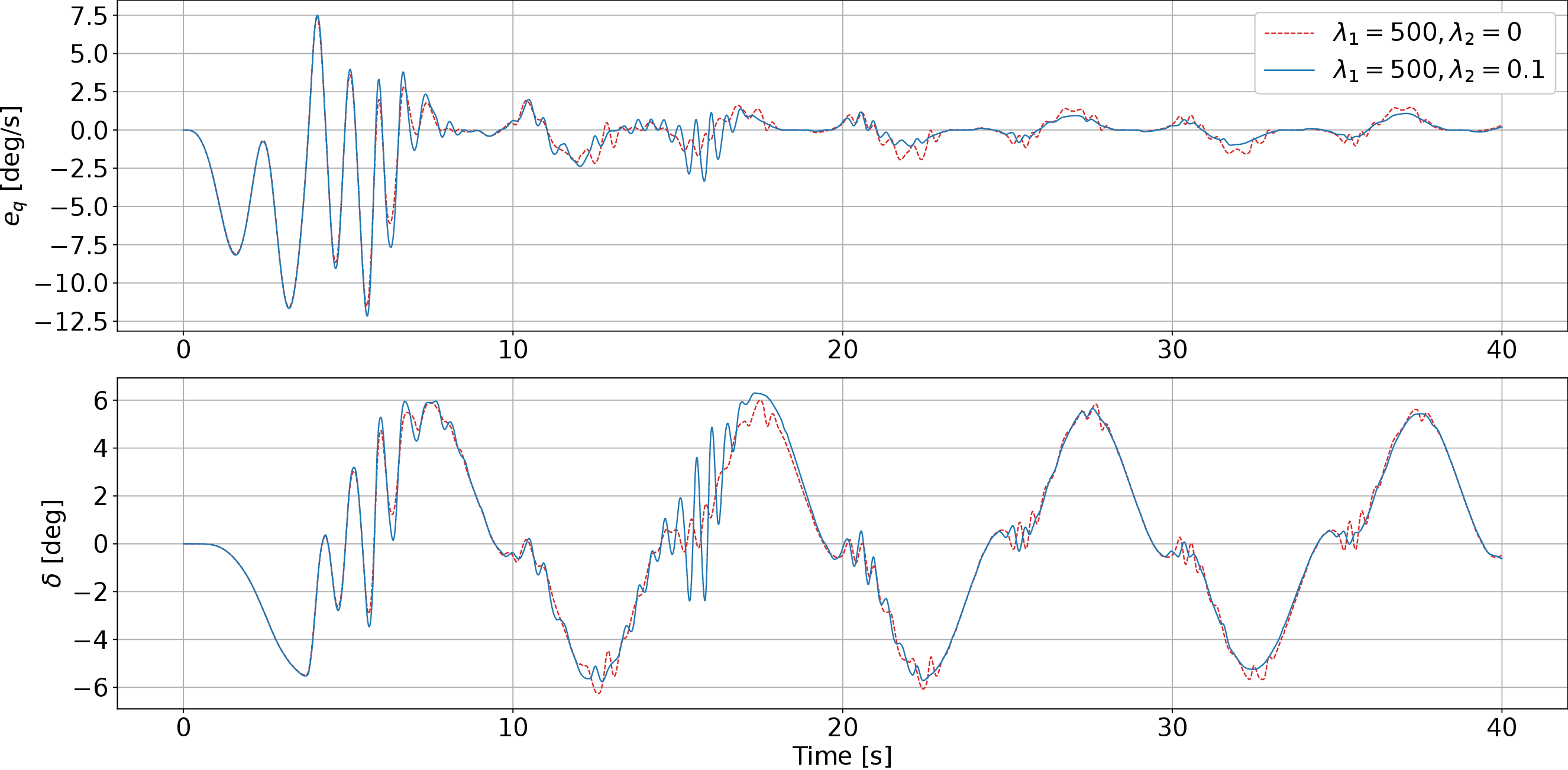}
    \caption{Comparison of pitch rate tracking error and control surface deflection between IHDP and Lyapunov function-guided IHDP for the lower-level agent.} \label{action_Cha4}
\end{figure}

Figure \ref{q_tracking_Cha4} compares pitch-rate tracking of two flight control systems. The first control system uses IHDP with a convergence metric in the higher-level agent, labeled by $\lambda_{1}=500,\lambda_{2}=0$. The second control system uses IHDP with a convergence metric in both higher-level and lower-level agents, labeled by $\lambda_{1}=500,\lambda_{2}=0.1$. $\lambda_{2}$ is the weight of convergence metric for the lower-level agent's actor. The tracking for the filtered pitch rate reference $q^{\prime}_{\text{ref}}$ is successful for both control systems. A closer look at Figure \ref{q_tracking_Cha4} shows successful tracking for high-frequency pitch rate reference in time periods 5-8s, 15-16s. Meantime, the tracking error has reduced in time periods 15-16s and 25-26s compared to that in time period 5-8s. This phenomena is also reflected in the first subplot of Figure \ref{action_Cha4}, indicating the control gains have grown sufficiently to handle system dynamics in the time period 0-10s.

Figure \ref{action_Cha4} compares tracking error $e_{q}$ and control surface deflection $\delta$. A phenomena is that $\delta$ exhibits high-frequency oscillations in the time period 15-17s. This is a result of the actor responding to the oscillatory pitch rate reference. A noteworthy observation from the comparison is that the case with $\lambda_{2} = 0.1$ achieves smoother and smaller tracking error than the case with $\lambda_{2} = 0$ during the stable phase (20–40s).

\begin{figure}[htbp]
    \centering
    \includegraphics[width=1.0\linewidth]{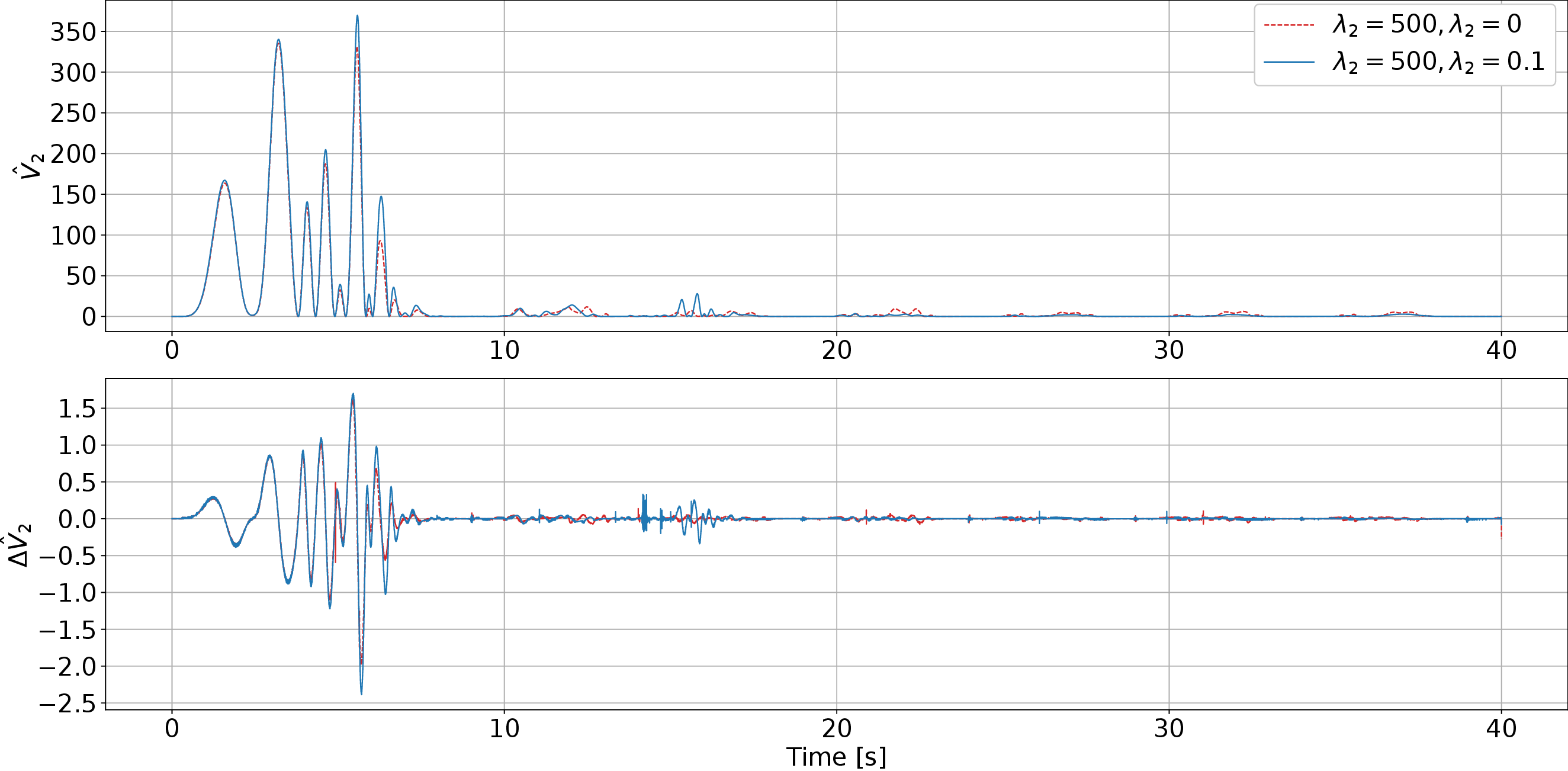}
    \caption{Comparison of critic output and its increment for the lower-level agent.} \label{q_Lyapunov_Cha4}
\end{figure}

Figure \ref{q_Lyapunov_Cha4} compares the Lyapunov function candidate $\hat{V}_{2}$. $\hat{V}_{2}$ increases and decreases rapidly during the starting learning phase 0-8s. The increases indicate the tracking error $e_{q(t)}$ is leaving the expected equilibrium point $e_{q}=0$. This results from the control gains being insufficient to offset the term $\dot{q}_{\text{ref}}$ and other dynamic terms in Equation \ref{aerialvehiclemodel}. The control gains eventually grow and diminish $\hat{V}_{2(t)}$. The second subplot compares the increment of Lyapunov function candidate defined by $\Delta \hat{V}_{2(t)}=\hat{V}_{2(t+1)}-\hat{V}_{2(t)}$. This comparison is less straightforward as the pitch rate references generated from the higher-level agents are slightly different. For example, the first control system ($\lambda_{2}=0$) generates a higher pitch rate reference in time periods 6-8s, leading to a larger control surface deflection. As a result, $\hat{V}_{2}$ in the case with $\lambda_{2}=0.1$ also shows a higher peak than the case with $\lambda_{2}=0$.





\section{Conclusion}\label{section_conclusion}
 
In this chapter, the convergence of the cascaded online learning flight control system is investigated. A convergence metric is designed based on the asymptotic stability condition, which requires that the Lyapunov function candidate decreases over time. The derivation of this metric accounts for both the incremental model approximation error and the state space discretization error. This convergence metric is incorporated into the actor's loss function using the Lagrangian method, enabling the actor to learn a control policy that explicitly considers convergence behavior. 

Simulation results demonstrate marginal improvements of decreases in the Lyapunov function candidate $\hat{V}_{1}$. The reasons lie in two aspects: (1) the higher-level actor uses the temporal smoothness losses, which penalizes increasing actions that help to lower the convergence metric. This reflects a performance \textit{trade-off} between action smoothness and convergence. (2) The tuning of the weight $\lambda_{1}$ is insensitive to the convergence performance, especially when the tracking error approaches zero. In this situation, the gradients from the Lyapunov function losses become very small and therefore do not exhibit clear improvements on tracking errors. On the other hand, the overweight of Lyapunov function loss may lead to the policy crossing the optimum which inversely increases this loss, especially when the tracking error approaches zero.

The comparison of $\hat{V}_{2}$ between the two control systems is not straightforward, as the pitch rate references differ slightly. However, the comparison of tracking error $e_q$ indicates that the control system using the convergence loss in the lower-level agent achieves smoother and smaller tracking errors than the one without it. Therefore, we recommend using same pitch rate reference to compared the lower-level actors in the future work.





\appendix

\subsection{Derivation of incremental model}\label{appendix_A}

Taking the Taylor expansion of systems \ref{aerialvehiclemodel}:

\begin{equation}\label{taylor_expanded_model}
\begin{aligned}
    \alpha_{t+1} =& \alpha_{t} + F^{1}_{t-1}(\alpha_{t}-\alpha_{t-1}) + G^{1}_{t-1}(q_{t}-q_{t-1})\\
    &+O\left[(\alpha_{t}-\alpha_{t-1})^{2},(q_{t}-q_{t-1})^{2}\right]\\
    q_{t+1} =& q_{t} + F^{2}_{t-1}(q_{t}-q_{t-1}) + G^{2}_{t-1}(\delta_{t}-\delta_{t-1})\\
    &+O\left[(q_{t}-q_{t-1})^{2},(\delta_{t}-\delta_{t-1})^{2}\right]\\
\end{aligned}
\end{equation}\\
where

\begin{equation}
\begin{aligned}
    &F^{1}_{t-1} = \frac{\partial h_{1} [\alpha_{t},\delta_{t},q_{t}]}{\partial \alpha_{t}}|_{\alpha_{t-1},\delta_{t-1},q_{t-1}}\\
    &G^{1}_{t-1} = \frac{\partial h_{1} [\alpha_{t},\delta_{t},q_{t}]}{\partial q_{t}}|_{\alpha_{t-1},\delta_{t-1},q_{t-1}}\\
    &F^{2}_{t-1} = \frac{\partial h_{2} [\alpha_{t},\delta_{t},q_{t}]}{\partial q_{t}}|_{\alpha_{t-1},\delta_{t-1},q_{t-1}}\\
    &G^{2}_{t-1} = \frac{\partial h_{2} [\alpha_{t},\delta_{t},q_{t}]}{\partial q_{t}}|_{\alpha_{t-1},\delta_{t-1},q_{t-1}}\\
\end{aligned}    
\end{equation}\\
and $O(\cdot)$ represents higher-order terms that can be ignored if the time step $t$ is sufficiently small.

Define state increments and control increment as
\begin{equation}
\begin{aligned}
    \Delta \alpha_{t} &= \alpha_{t}-\alpha_{t-1}\\ \Delta q_{t} &= q_{t} -q_{t-1}\\
    \Delta \delta_{t}&=\delta_{t}-\delta_{t-1}
\end{aligned}    
\end{equation}

Therefore, the incremental model can be formulated as
\begin{equation}
\begin{aligned}
    &\Delta \alpha_{t+1} = F^{1}_{t-1}\Delta\alpha_{t}+G^{1}_{t-1}\Delta q_{t}\\
    &\Delta q_{t+1} = F^{2}_{t-1}\Delta q_{t}+G^{2}_{t-1}\Delta \delta_{t}
\end{aligned}
\end{equation}

\subsection{Higher-level agent's critic and actor update}\label{appendix_B}

\subsubsection{Critic}
The gradient of $L^{C_{1}}_{t}$ with respect to $\psi_{1}$ is
\begin{equation}
\begin{aligned}
     \frac{\partial L^{C_{1}}_{t}}{\partial \psi_{1(t)}} &= \frac{\partial L^{C_{1}}_{t}}{\partial \delta_{1(t)}} \frac{\partial \delta_{1(t)}}{\partial \psi_{1(t)}}\\
     &=-\delta_{1(t)} \frac{\partial \hat{V}_{1}(\alpha_{t},e_{\alpha(t)})}{\partial \psi_{1(t)}}
\end{aligned}
\end{equation}

The parameter set $\psi_{1}$ is updated as

\begin{equation}
        \psi_{1(t+1)} = \psi_{1(t)} - \eta_{C_{1}}\frac{\partial L^{C_{1}}_{t}}{\partial \psi_{1(t)}}
\end{equation}\\
where $\eta_{C_{1}}$ is the learning rate.

\subsubsection{Target Critic}
The target critic network is used to stabilize learning by delaying the updates:

\begin{equation}\label{target_critic_1_update}
    \psi^{\prime}_{1(t+1)} = \tau \psi_{1(t+1)} + (1-\tau)\psi^{\prime}_{1(t)}
\end{equation}\\
where $\tau$ is delay factor $\tau$, $\psi^{\prime}_{1}$ is the parameter set of target critic.

\subsubsection{Actor}
The gradient of $L^{A_{1}}_{t}$ with respect to $\vartheta_{1}$ is

\begin{equation}\label{loss_actor_1_gradient}
\begin{aligned}
    \frac{\partial L^{A_{1}}_{t}}{\partial \vartheta_{1(t)}} 
     =& \frac{\partial \left[c_{1}(\hat{e}_{\alpha(t+1)},q_{t})+\gamma \hat{V}_{1}(\hat{\alpha}_{t+1},\hat{e}_{\alpha(t+1)})\right]}{\partial  \vartheta_{1(t)}}\\ 
    = &\left[\frac{\partial c_{1}(\hat{e}_{\alpha(t+1)},q_{t})}{\partial \hat{\alpha}_{t+1}} + \gamma\frac{\hat{V}_{1}(\hat{\alpha}_{t+1},\hat{e}_{\alpha(t+1)})}{\partial \hat{\alpha}_{t+1}}\right]\\
    &\;\; \frac{\partial \hat{\alpha}_{t+1}}{\partial q_{\text{ref}}(t)} \frac{\partial q_{\text{ref}}(t)}{\partial  \vartheta_{1(t)}}\\
    =& \left[\frac{\partial c_{1}(\hat{e}_{\alpha(t+1)},q_{t})}{\partial \hat{\alpha}_{t+1}} + \gamma\frac{\hat{V}_{1}(\hat{\alpha}_{t+1},\hat{e}_{\alpha(t+1)})}{\partial \hat{\alpha}_{t+1}}\right]\\
    &\;\; \hat{G}^{1}_{t-1} \frac{\partial q_{\text{ref}}(t)}{\partial  \vartheta_{1(t)}}
\end{aligned}
\end{equation}

The parameter set $\vartheta_{1}$ is updated as

\begin{equation}
    \vartheta_{1(t+1)} = \vartheta_{1(t)} -\eta_{A_{1}} \frac{\partial L^{A1}_{t}}{\partial \vartheta_{1(t)}} 
\end{equation}\\
where $\eta_{A_{1}}$ is the learning rate.

From an algorithmic perspective, the update of $\psi_{t}$ to $\psi_{t+1}$ can be performed multiple times until the critic loss $L_{\text{critic}}$ falls below a specified threshold. Subsequently, the update of $\vartheta_{t}$ to $\vartheta_{t+1}$ can also be repeated until the actor loss $L_{\text{actor}}$ begins to reverse direction, indicating that $\vartheta_{t}$ is near a local optimum. These steps constitute one iteration of approximate value iteration. The approximate value iteration can be executed for multiple times at time step $t$.

\subsection{Lower-level agent's critic and actor update}\label{appendix_C}

\subsubsection{Critic}

The gradient of $L^{C_{2}}_{t}$ with respect to $\psi_{2}$ is
\begin{equation}
\begin{aligned}
    \frac{\partial L^{C_{2}}_{t}}{\partial \psi_{2}(t)}=& \frac{\partial L^{C_{2}}_{t}}{\partial \delta_{2(t)}} \frac{\partial \delta_{2(t)}}{\partial \psi_{2}(t)}\\
    =&-\delta_{2(t)} \frac{\partial \hat{V}(q_{t},e_{q(t)})}{\partial \psi_{2}(t)}
\end{aligned}
\end{equation}

The parameter set $\psi_{2}$ is updated as
\begin{equation}
    \psi_{2}(t+1) = \psi_{2}(t) - \eta_{C_{2}}\frac{\partial L^{C_{2}}_{t}}{\partial \psi_{2}(t)}
\end{equation}\\
where $\eta_{C_{2}}$ is the learning rate.

\subsubsection{Target Critic}
Target critic is used to stabilize the learning by slowing down the network update.
\begin{equation}
    \psi^{\prime}_{2(t+1)} = \tau \psi_{2(t+1)} + (1-\tau)\psi^{\prime}_{2(t)}
\end{equation}\\
where $\tau$ is delay factor, $\psi^{\prime}_{2}$ is parameter set of target critic.

\subsubsection{Actor}

The gradient of $L^{A2}_{t}$ with respect to $\vartheta_{2}$ is

\begin{equation}
\begin{aligned}
    \frac{\partial L^{A2}_{t}}{\partial \vartheta_{2(t)}} 
    =& \frac{\partial \left[c_{2}(\hat{e}_{q(t+1)},\delta_{t})+\gamma \hat{V}_{2\text{target}}(\hat{q}_{t+1},\hat{e}_{q(t+1)})\right]}{\partial  \vartheta_{2(t)}}\\ 
    =& \left[\frac{\partial c_{2}(\hat{e}_{q(t+1)},\delta_{t})}{\partial \hat{q}_{t+1}} + \gamma\frac{\hat{V}_{2\text{target}}(\hat{q}_{t+1},\hat{e}_{q(t+1)})}{\partial \hat{q}_{t+1}}\right]\\
    &\;\;\frac{\partial \hat{q}_{t+1}}{\partial \delta_{e(t)}} \frac{\partial \delta_{e(t)} }{\partial  \vartheta_{2(t)}}\\
    =& \left[\frac{\partial c_{2}(\hat{e}_{q(t+1)},\delta_{t})}{\partial \hat{q}_{t+1}} + \gamma\frac{\hat{V}_{2\text{target}}(\hat{q}_{t+1},\hat{e}_{q(t+1)})}{\partial \hat{q}_{t+1}}\right]
    \\
    &\;\;\;\hat{G}^{2}_{t-1} \frac{\partial \delta_{e(t)} }{\partial  \vartheta_{2(t)}}\\
\end{aligned}
\end{equation}

The parameter set $\vartheta_{2}$ is updated as

\begin{equation}
    \vartheta_{2(t+1)} = \vartheta_{2(t)} -\eta_{A_{2}} \frac{\partial L^{A2}_{t}}{\partial \vartheta_{2(t)}} 
\end{equation}\\
where $\eta_{A_{2}}$ is the learning rate.

\end{document}